\documentclass[letterpaper, 10 pt, conference]{ieeeconf}  % Comment this line out if you need a4paper

\IEEEoverridecommandlockouts                              % This command is only needed if 
\usepackage{graphics} % for pdf, bitmapped graphics files
\usepackage{epsfig} % for postscript graphics files
\usepackage{mathptmx} % assumes new font selection scheme installed
\usepackage{times} % assumes new font selection scheme installed
\usepackage{amsmath} % assumes amsmath package installed
\usepackage{amssymb}  % assumes amsmath package installed

\usepackage{hyperref}
\usepackage{url}
\usepackage{graphicx}
\usepackage{xcolor}
\usepackage{wrapfig}
\usepackage{subcaption}

\newcommand{\cmt}[1]{}

\newcommand{\algname}{{PrivilegedDreamer\ }}

\long\def\ignorethis#1{}

\newcommand{\pctab}{\hspace{0.2in}}

\title{\LARGE \bf
PrivilegedDreamer: Explicit Imagination of Privileged Information
\\
for Rapid Adaptation of Learned Policies
}

\author{Morgan Byrd$^{1*}$, Jackson Crandell$^{1}$, Mili Das$^{1}$, Jessica Inman$^{2}$, Robert Wright$^{2}$, and Sehoon Ha$^{1}$% <-this % stops a space
\thanks{This work was supported by the GTRI Graduate Student Researcher Fellowship Program and NSF Award No. 2339076}% <-this % stops a space
\thanks{$^{1}$Georgia Institute of Technology, Atlanta, GA, 30308, USA}%
\thanks{$^{2}$Georgia Tech Research Institute, Atlanta, GA, 30308, USA}%
\thanks{*Correspondence to abyrd45@gatech.edu}
}

\begin{document}

\maketitle
\thispagestyle{empty}
\pagestyle{empty}

%%%%%%%%%%%%%%%%%%%%%%%%%%%%%%%%%%%%%%%%%%%%%%%%%%%%%%%%%%%%%%%%%%%%%%%%%%%%%%%%
\begin{abstract}
Numerous real-world control problems involve dynamics and objectives affected by unobservable hidden parameters, ranging from autonomous driving to robotic manipulation, which cause performance degradation during sim-to-real transfer. To represent these kinds of domains, we adopt hidden-parameter Markov decision processes (HIP-MDPs), which model sequential decision problems where hidden variables parameterize transition and reward functions. Existing approaches, such as domain randomization, domain adaptation, and meta-learning, simply treat the effect of hidden parameters as additional variance and often struggle to effectively handle HIP-MDP problems, especially when the rewards are parameterized by hidden variables. We introduce PrivilegedDreamer, a model-based reinforcement learning framework that extends the existing model-based approach by incorporating an explicit parameter estimation module. PrivilegedDreamer features its novel dual recurrent architecture that explicitly estimates hidden parameters from limited historical data and enables us to condition the model, actor, and critic networks on these estimated parameters. Our empirical analysis on five diverse HIP-MDP tasks demonstrates that PrivilegedDreamer outperforms state-of-the-art model-based, model-free, and domain adaptation learning algorithms. Additionally, we conduct ablation studies to justify the inclusion of each component in the proposed architecture.  
\end{abstract}

\section{Introduction}

Markov decision processes (MDPs) have been powerful mathematical frameworks for modeling a spectrum of sequential decision scenarios, from computer games to intricate autonomous driving systems; however, they often assume fixed transition or reward functions. In many real-world domains, there exist hidden-parameter MDPs (HIP-MDPs)~\cite{Doshi-Velez2016} that are characterized by the presence of hidden or uncertain parameters playing significant roles in their dynamics or reward functions. For instance, autonomous driving must handle diverse vehicles with distinctive dynamic attributes and properties to achieve a better driving experience, while the agricultural industry must account for variations in produce weight for sorting. Consequently, research endeavors have explored diverse approaches, including domain randomization \cite{Tobin2017}, domain adaptation \cite{Peng2020}, and meta-learning \cite{Wang2016}, to address these challenges effectively. 

We approach these HIP-MDP problems using model-based reinforcement learning because a world model holds significant promise in efficiently capturing dynamic behaviors characterized by hidden parameters, ultimately resulting in improved policy learning. Particularly, we establish our framework based on Dreamer \cite{Hafner2019}, which has been effective in solving multiple classes of problems, including DeepMind Control Suite \cite{Tassa2018}, Atari \cite{Hafner2020}, and robotic control \cite{Wu2022}. Our initial hypothesis was that the Dreamer framework could capture parameterized dynamics accurately by conditioning the model on latent variables, leading to better performance at the end of learning. 
However, Dreamer is designed to predict action-conditioned dynamics in the observation space and does not consider the effect of hidden parameters.
% However, our initial experiments reveal that Dreamer often learns suboptimal control policies because it does not have enough motivation to dream about hidden parameters while merely trying to improve the averaged performance. 

This paper presents \algname to solve HIP-MDPs via explicit prediction of hidden parameters. Our key intuition is that a recurrent state-space model (RSSM) of model-based RL must be explicitly conditioned on hidden parameters to capture the subtle changes in dynamics or rewards. However, a HIP-MDP assumes that hidden variables are not available to agents. Therefore, we introduce an explicit module to estimate hidden parameters from a history of state variables via a long short-term memory (LSTM) network, which can be effectively trained by minimizing an additional reconstruction loss. This dual recurrent architecture allows accurate estimation of hidden parameters from a short amount of history. The estimated hidden parameters are also fed into the transition model, actor, and critic networks to condition their adaptive behaviors on these hidden parameters. 
% This paper presents \algname to solve HIP-MDP via explicit imagination of hidden parameters. Our \algname incorporates two technical novelties: concurrent state estimation and privileged learning. The first technical component is an estimation module that predicts hidden parameters from state variables, inspired by concurrent state estimation and policy learning~\citep{Ji2022}. However, simply adding a reconstruction loss to the recurrent state variable is insufficient for accurate estimation. Instead, we introduce a separate Long short-term memory (LSTM) network solely dedicated to state estimation. We then feed the estimated hidden parameters to the policy network. Second, we use the ground-truth hidden parameters for the training of the policy and value networks, inspired by teaching agent training in \emph{learning by cheating} philosophy [Learning by Cheat, DreamWaQ]. This allows the framework to efficiently learn all the components, including model, actor, and critic networks, without divergence. 

We evaluate our method in five HIP-MDP environments, two of which also have hidden-parameter-conditioned reward functions. We compare our method against several state-of-the-art baselines, including model-based (DreamerV2~\cite{Hafner2020}), model-free (Soft Actor Critic~\cite{Haarnoja2018} and Proximal Policy Optimization~\cite{Schulman2017}), and domain adaptation (Rapid Motor Adaptation~\cite{Kumar2021}) algorithms. Our \algname achieves a $41$\% higher average reward over these five tasks and demonstrates remarkably better performance on the HIP-MDPs with parameterized reward functions. We further analyze the behaviors of the learned policies to investigate how rapid estimation of hidden parameters affects the final performance and also to justify the design decisions of the framework. Finally, we outline a few interesting future research directions.

\section{Related Work}

\subsection{World Models}

Model-based RL improves sample efficiency over model-free RL by learning an approximate model for the transition dynamics of the environment, allowing for policy training without interacting with the environment itself. However, obtaining accurate world models is not straightforward because the learned model can easily accumulate errors exponentially over time. To alleviate this issue, \cite{Chua2018} designs ensembles of stochastic dynamics models to attempt to incorporate uncertainty. The Dreamer architecture~\cite{Hafner2019,Hafner2020,Hafner2023} learns a model of the environment via reconstructing the input from a latent space using the recurrent state-space model (RSSM). The RSSM incorporates the Gated Recurrent Unit (GRU) network~\cite{Cho2014} and the Variational Autoencoder (VAE)~\cite{Kingma2013} for modeling. With this generative world model, the policy is trained with imagined trajectories in this learned latent space.  \cite{Robine2023} and \cite{Micheli2022} leverage the Transformer architecture~\cite{Vaswani2017} to autoregressively model the world dynamics and similarly train the policy in latent imagination. 
Our work is built on top of the Dreamer architecture, but the idea of explicit modeling of hidden parameters has the potential to be combined with other architectures.
% \sehoon{Morgan, try to avoid some vauge words, like ``use''. It is better to be specific. It is also better to avoid the repetition of the same words}

\subsection{Randomized Approaches without Explicit Modeling}
One of the most popular approaches to deal with uncertain or parameterized dynamics is domain randomization (DR), which aims to improve the robustness of a policy by exposing the agent to randomized environments. It has been effective in many applications, including manipulation \cite{Peng2018, Tobin2017, Zhang2016, James2017}, locomotion \cite{Peng2020, Tan2018}, autonomous driving \cite{Tremblay2018}, and indoor drone flying \cite{Sadeghi2016}. 
% \sehoon{assign citations to the proper application category}. 
Domain randomization has also shown great success in deploying trained policies on actual robots, such as performing sim-to-real transfer for a quadrupedal robot in \cite{Tan2018} and improving performance for a robotic manipulator in \cite{Peng2018}. \cite{rigter2024waker} and \cite{yamada2023twist} both incorporate DR for increased robustness in a world model setting. While DR is highly effective in many situations, it tends to lead to an overly conservative policy that is suboptimal for challenging problems with a wide range of transition or reward functions. 

\subsection{Domain Adaptation}
Since incorporating additional observations is often beneficial \cite{Kim2020ObservationSM}, another common strategy for dealing with variable environments is to incorporate the hidden environmental parameters into the policy for adaptation. This privileged information of the hidden parameters can be exploited during training, but at test time, system identification must occur online. For model-free RL, researchers typically train a universal policy conditioned on hidden parameters and estimate them at test time by identifying directly from a history of observations~\cite{Yu2017,Kumar2021,Nahrendra2023}. Another option is to improve state estimation while training in diverse environments, which similarly allows for adaptation without needing to perform explicit system identification~\cite{Ji2022}. 
% \sehoon{Morgan, check the descriptions and adjust them.} 
For model-based RL, the problem of handling variable physics conditions is handled in multiple ways. A few research groups \cite{Nagabandi2018,Sæmundsson2018} propose using meta-learning to rapidly adapt to environmental changes online. \cite{Wang2021} uses a graph-based meta RL technique to handle changing dynamics. \cite{Ball2021} used data augmentation in offline RL to get zero-shot dynamics generalization. The most applicable methods for our work are those that use a learned encoder to estimate a context vector that attempts to capture the environmental information. Then, this context vector is used for conditioning the policy and forward prediction, as in \cite{Wang2022,Lee2020,Huang2021,Seo2020}.

% Another common choice for dealing with a variable environment is to attempt to incorporate environmental information into the policy for adaptation. During training, privileged information can be used, but at test time, system identification must occur online.

% For model-free RL, \cite{Yu2017} and \cite{Kumar2021} both use privileged environmental information to learn a universal policy during training and then use a history of observations to train an encoder to estimate the parameters during testing. \cite{Ji2022} simultaneously trains a policy and estimation network in a changing environment to improve performance and avoid explicitly having to perform system identification.  \cite{Nahrendra2023} similarly employs an observation history to estimate velocity and as an input to a VAE that attempts to reconstruct the privileged state information.

% For model-based RL, the problem of handling variable physics conditions is handled in multiple ways. \cite{Nagabandi2018} and \cite{Sæmundsson2018} propose using meta-learning to rapidly adapt to dynamics changes online. \cite{Wang2021} uses a graph-based meta RL technique to handle changing dynamics. \cite{Ball2021} used data augmentation in offline RL to get zero-shot dynamics generalization. The most applicable methods for our work are the problems that use a learned encoder to estimate a context vector that attempts to capture the environmental information and is used to condition the policy and for forward prediction, as in \cite{Wang2022}, \cite{Lee2020}, \cite{Huang2021}, \cite{Seo2020}.

\section{PrivilegedDreamer: Adaptation via Explicit Imagination}

\subsection{Background}
\subsubsection{Hidden-parameter MDP}
% Some background information - Hidden parameter MDP setup
% \cite{Doshi-Velez2016}
% Clean up the notation and add citations
A Markov decision process (MDP) formalizes a sequential decision problem, which is defined as a tuple $(S, A, T, R, p_0)$, where $S$ is the state space, $A$ is the action space, $T$ is the transition function, $R$ is the reward function, and $p_0$ is the initial state distribution. For our work, we consider the hidden-parameter MDP (HIP-MDP), which generalizes the MDP by conditioning the transition function $T$ and/or the reward function $R$ on an additional hidden latent variable $\omega$ sampled from a distribution $p_\omega$ \cite{Doshi-Velez2016}. Without losing generality, $\omega$ can be a scalar or a vector. In the setting of continuous control, which is the primary focus of this work, this latent variable represents physical quantities, such as mass or friction, that govern the dynamics but are not observable in the state space.

% Some description of Dreamer setup/ actor-critic learning
% Needs description of variables and improved layout
\subsubsection{Dreamer}
For our model, we build upon the DreamerV2 model of \cite{Hafner2020}. DreamerV2 uses a recurrent state-space model (RSSM) to model dynamics and rewards. This RSSM takes as input the state $x_t$ and the action $a_t$ to compute a deterministic recurrent state $h_t = f_\phi(h_{t-1}, z_{t-1}, a_{t-1})$ using a GRU $f_\phi$ and a sampled stochastic state $z_t \sim q_\phi(z_t \vert h_t, x_t)$ using an encoder $q_\phi$. The transition predictor $\hat{z}_t \sim p_\phi(\hat{z}_t \vert h_t)$ computes an estimate $\hat{z}_t$ of the stochastic state using only the deterministic state $h_t$, which is necessary during training in imagination as $x_t$ is not available. The combination of the deterministic and stochastic states is used as a representation to reconstruct the state $\hat{x}_t \sim p_\phi(\hat{x}_t \vert h_t, z_t)$, predict the reward $\hat{r}_t \sim p_\phi(\hat{r}_t \vert h_t, z_t)$, and predict the discount factor $\hat{\gamma}_t \sim p_\phi(\hat{\gamma}_t \vert h_t, z_t)$. 

% \begin{align*}
%     \text{Recurrent model: } h_t &= f_\phi(h_{t-1}, z_{t-1}, a_{t-1}) \notag\\
%     \text{Representation model: } z_t &\sim q_\phi(z_t \vert h_t, x_t) \notag\\
%     \text{Transition predictor: } \hat{z}_t &\sim p_\phi(\hat{z}_t \vert h_t) \notag\\
%     \text{Input predictor: } \hat{x}_t &\sim p_\phi(\hat{x}_t \vert h_t, z_t) \notag\\
%     \text{Reward predictor: } \hat{r}_t &\sim p_\phi(\hat{r}_t \vert h_t, z_t) \notag\\
%     \text{Discount predictor: } \hat{\gamma}_t &\sim p_\phi(\hat{\gamma}_t \vert h_t, z_t) \notag\\
% \end{align*}

For policy learning, Dreamer adopts an actor-critic network, which is trained via imagined rollouts. For each imagination step $t$, the latent variable $\hat{z}_t$ is predicted using only the world model, the action is sampled from the stochastic actor: $a_t \sim \pi_\theta(a_t | \hat{z}_t)$, and the value function is estimated as: $v_\psi \approx \mathbb{E}_{p_\phi, p_\theta} [\sum \gamma^{\tau-t}\hat{r}_\tau]$, where $\hat{r}_t$ is computed from the reward predictor above. The actor is trained to maximize predicted discounted rewards over a fixed time horizon $H$. The critic aims to accurately predict the value from a given latent state. The actor and critic losses are: 

\vspace{-1em}

\begin{align*}
    L_{actor} &=  \mathbb{E}_{p_\phi, p_\theta} \Bigg[\sum_{t=1}^{H-1} -V_t^\lambda- \eta H[a_t|\hat{z}_t] \Bigg] \\
    L_{critic} &= \mathbb{E}_{p_\phi, p_\theta} \Bigg[\sum_{t=1}^{H-1} \frac{1}{2}\left(v_\psi(\hat{z}_t) - \text{sg}(V_t^\lambda)\right)^2 \Bigg]
\end{align*}

% Algorithm description, differences from Dreamer architecture
% Figure/algorithm

\subsection{Algorithm}
% \paragraph{PrivilegedDreamer}
While the original DreamerV2 layout works effectively for many tasks, it falters in the HIP-MDP domain, especially in the case where the reward explicitly depends on the hidden latent variable. Even though the RSSM has memory to determine the underlying dynamics, prior works, such as \cite{Seo2020}, have shown that this hidden state information is poorly captured implicitly and must be learned explicitly. 

\subsubsection{Explicit Imagination via LSTM} 
% While this setup gives a performance improvement, it is still lacking as the privileged information $\omega$ is only estimated by the generative model of the RSSM but is not an input to the world model. This limits the effectiveness of the policy training as the latent features $z_t$ do not fully capture the effect of this information from the additional reconstruction term alone. Thus, predictions via imagination are inaccurate, leading to suboptimal performance.

% Maybe want a figure of this external module
To improve the estimation of hidden parameters, we incorporate an additional independent module for estimating the privileged information from the available state information. This dual recurrent architecture allows us to effectively estimate the important hidden parameters in the first layer and model other variables conditioned on this estimation in the second layer. Our estimation module $\tilde{\omega}_t \sim \eta_\phi(\tilde{\omega}_t \vert x_t, a_{t-1})$ takes the state $x_t$ and previous action $a_{t-1}$ as inputs and predicts the intermediate hidden parameter $\tilde\omega_t$. It is still parameterized by $\phi$ because we treat it as part of the world model.
% \sehoon{I introduced $\eta$ for highlighting this estimation network. Please check the consistency.} 
 The estimation module is comprised of an LSTM~\cite{Hochreiter} followed by MLP layers that reshape the output to that of the privileged data. We use an LSTM because its recurrent architecture is more suitable to model subtle and non-linear relationships between state and hidden variables over time. However, the choice of the architecture was not significant to the performance. In our experience, LSTM and GRU demonstrated similar performance. 
% It is trained along with the world model, and the current estimate $\omega_{est}$ is used as a world model input, although the gradient is stopped.

Note that we use $\tilde\omega_t$ to make the recurrent world model conditioned on the estimated hidden variable. For the actor and critic, we feed the value from the prediction head, $\hat\omega_t$, which will be described in the next paragraph.

\subsubsection{Improving Accuracy via Additional Prediction Head}
We also added an additional prediction head $p_{\phi} (\hat\omega_t \vert h_t, z_t)$, which is similar to the reward or state prediction heads. While the previous LSTM estimation $\eta$ predicts the intermediate parameter $\tilde\omega_t$ to make the model conditioned on the hidden parameter, this additional prediction head offers two major improvements: 1) encouraging the RSSM state variables $h_t$ and $z_t$ to contain enough information about the hidden parameter and 2) improving the prediction accuracy.

\subsubsection{Hidden Variable Loss}
We design an additional loss to train the estimation module, which is similar to the other losses of the DreamerV2 architecture. We do not use the discount predictor from the original DreamerV2 architecture as all of our tests are done in environments with no early termination. We group the other Dreamer losses all under $L_{Dreamer}$ to highlight our differences. This makes the total loss for the world model:
\vspace{-0.75em}
\begin{multline}
    L(\phi) = L_{Dreamer} + 
    \mathbb{E}_{q_{\phi}(z_{1:T} \vert a_{1:T},x_{1:T},\omega_{1:T})} \\ \Bigg[ \sum_{t=1}^T - \ln \eta_{\phi} (\tilde{\omega}_t \vert x_t, a_{t-1}) - \ln p_{\phi} (\hat{\omega}_t \vert h_t, z_t) \Bigg].
\end{multline}
% \end{equation*}

where the first additional loss term is to compute an intermediate estimate $\tilde{\omega}$ for the hidden parameter $\omega$ using the environment states $x$ and actions $a$ and the second term is the world model reconstruction loss for $\hat\omega$ based on the RSSM latent variables $h$ and $z$.

% \begin{equation*}
% \begin{aligned}
%     L(\phi) = \E_{q_{\phi}(z_{1:T} \vert a_{1:T},x_{1:T},\omega_{1:T})}[\sum_{t=1}^T &-\ln p_{\phi}(x_t \vert h_t, z_t) - \ln p_{\phi} (r_t \vert h_t, z_t) - \underbrace{\ln \eta_{\phi} (\omega_t \vert h_t, z_t)}_{\textit{hidden variable loss}}\\ &+ \beta \text{KL}[q_{\phi} (z_t \vert h_t, x_t, \omega_t) || p_{\phi} (z_t \vert h_t)]].
% \end{aligned}
% \end{equation*}

It is important to highlight that relying solely on this hidden parameter loss term is not sufficient. Theoretically, it seems like the loss encourages the recurrent state variables $h_t$ and $z_t$ to encapsulate all relevant information and increase all the model, actor, and critic networks' awareness of hidden parameters. However, in practice, this privileged information remains somewhat indirect to those networks. Consequently, this indirect access hinders their ability to capture subtle changes and results in suboptimal performance.

\begin{figure}
% \vspace{-2em}
\begin{center}
\includegraphics[width=\linewidth]{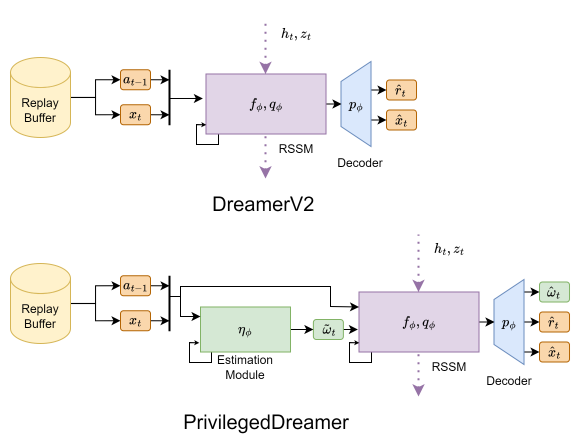}
\end{center}
\caption{Architecture of PrivilegedDreamer. Compared to the default DreamerV2 model (\textbf{top}), our architecture (\textbf{bottom}) adopts an explicit parameter estimation model $\eta$ to predict the hidden parameters $\omega_t$ from a history of states. Then, the estimated parameters $\tilde{\omega}_t$ are fed into the model to establish the explicit dependency.}
% \caption{World model training architecture. We collect interactions with the environment into a replay buffer. We sample states $x_t$ which we use to estimate the hidden parameter $\omega_t$. These are combined with the recurrent deterministic state $h_t$ and are given as inputs to the encoder $q_\phi$ to generate the stochastic state $z_t$. This full latent state is then passed to a decoder to reconstruct the states $x_t$, rewards $r_t$, and hidden parameter $\omega_t$. Our addition of the hidden parameter $\omega_t$ to both the encoder and decoder leads to a world model representation that is much more capable of dealing with variations in $\omega_t$ than the default Dreamerv2 model. \sehoon{Also highlight our invention. What are some unnamed modules? (red and blue)} \sehoon{Many variables without explanation. How can we improve the readability?}} \morgan{Some adjustments, probably needs more work}
\label{fig:WM_model_architecture}
\end{figure}

% Something about quickly determining value of privileged information online
% for adaptation
\subsubsection{Hidden-parameter Conditioned Networks (ConditionedNet)}
Once we obtain the estimate of the hidden parameter $\omega_t$, we feed this information to the networks. This idea of explicit connection has been suggested in different works in reinforcement learning, such as rapid motor adaptation (RMA)~\cite{Kumar2021} or meta strategy optimization (MSO)~\cite{Yu2020}. Similarly, we augment the inputs of the representation model $z_t$, the critic network $v_\psi$, and the actor network $\pi_\theta$ to encourage them to incorporate the estimated $\tilde\omega_t$ and $\hat\omega_t$.

% Another improvement in performance comes from incorporating the input state information along with the hidden variable into the actor and critic networks. While the original Dreamerv2 used only the latent variable $z_t$ as input, we also use the hidden variable $\omega$ and the state $x_t$.
% The intuition for also incorporating $\omega_t$ in the policy input can be seen from works like RMA \citep{Kumar2021} or MSO \citep{Yu2020}, which show that directly incorporating environmental information as inputs give better results over other methods of handling changing environments, like domain randomization. 

\subsubsection{Additional Proprioceptive State as Inputs}
In our experience, it is beneficial to provide the estimated state information as additional inputs to the actor and critic networks. We hypothesize that this may be because the most recent state information $x_t$ is highly relevant for our continuous control tasks. On the other hand, the RSSM states $h_t$ and $z_t$ are indirect and more suitable for establishing long-term plans.
% In our experience, it is encouraged to estimate the proprioceptive state information $x_t$ also provided the performance gain for our proprioceptive control problems. We hypothesize that this may be the case because the most recent information is the most relevant for these continuous control tasks, and directly including the latest states better captures this short-term information than the RSSM features directly, as the recurrent structure must represent more long-term features. 
% \sehoon{Morgan, any intuition?} \morgan{Maybe something like this?}

% During policy training in imagination, we use the reconstructed states $\hat{x}_t$, which are already available from the world model loss, and the ground truth hidden variable $\omega$, which is obtained from the replay buffer which initializes the imagined rollout. At inference time, we use the actual state information and an estimate for $\omega$, obtained from the world model predictor. \sehoon{We may be able to connect the concept to MSO or RMA here} \morgan{Tried to connect to using privileged information like RMA}

% Same as with original Dreamer but include necessary changes to world model
% stuff

\subsubsection{Summary}
On top of DreamerV2, Our \algname includes the following components:

\vspace{-0.5em}

\begin{minipage}{0.35\textwidth}
    \scalebox{0.88}{ % Adjust the scaling factor as needed
        \parbox{\textwidth}{
            \begin{align*}
                &\text{Recurrent hidden parameter predictor:} & \tilde{\omega}_t &\sim \eta_\phi(\tilde{\omega}_t \vert h_t, z_t) \\ 
                &\text{HIP-conditioned representation model:} & z_t &\sim q_\phi(z_t \vert h_t, x_t, \tilde{\omega}_t) \\
                &\text{HIP prediction head:} & \hat{\omega}_t &\sim p_\phi(\hat{\omega}_t \vert h_t, z_t) \\
                &\text{HIP-conditioned critic:} & v_t &\sim v_\psi(v_t \vert h_t, z_t, x_t, \hat{\omega}_t) \\
                &\text{HIP-conditioned actor:} & \hat{a}_t &\sim \pi_\theta(a_t \vert h_t, z_t, x_t, \hat{\omega}_t)
            \end{align*}
        }
    }
\end{minipage}

\begin{figure*}
% \centering
    \centering
    \begin{tabular}{c c c c c}
        \includegraphics[height=3cm,trim={0.2cm 0.2cm 14cm 0.25cm},clip]{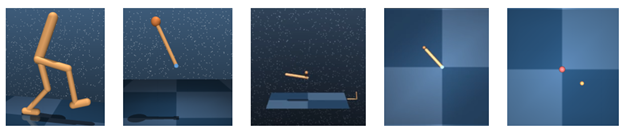} &
        \includegraphics[height=3cm,trim={3.3cm 0.2cm 10.4cm 0.25cm},clip]{images/Tasks.png} &
        \includegraphics[height=3cm,trim={6.7cm 0.2cm 6.8cm 0.25cm},clip]{images/Tasks.png} &
        \includegraphics[height=3cm]{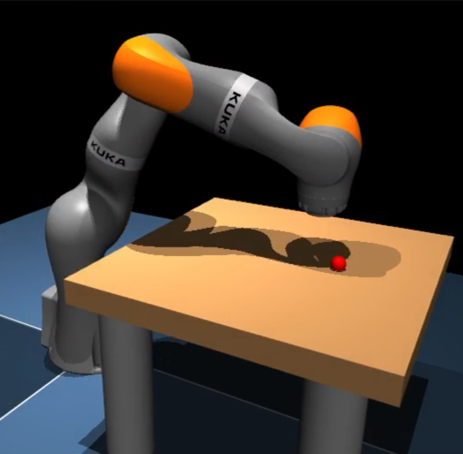} & 
        \includegraphics[height=3cm,trim={13.6cm 0.2cm 0.1cm 0.25cm},clip]{images/Tasks.png} \\
        Walker & Pendulum & Throwing & Kuka Sorting & Pointmass \\
    \end{tabular}
\caption{Five HIP-MDP tasks used in our experiments.}
\label{fig:tasks}
\end{figure*}
\vspace{-1em}

% \sehoon{Highlight the changes over Dreamer} \morgan{Highlighted in red}
\vspace{0.75em}
We omit the unchanged components from DreamerV2, such as input and reward predictors, for brevity.
% Our modifications over the original Dreamerv2 architecture are highlighted in red.
A schematic of the model architecture used for training the world model itself can be seen in Fig.~\ref{fig:WM_model_architecture}. This setup trains the encoder network, decoder network, and the latent feature components $z$ and $h$. The estimation module $\eta$ that initially estimates the value of $\tilde\omega_t$ is also trained here.

\renewcommand{\arraystretch}{1.2}
\begin{table*}[t]
\centering
\begin{tabular}[c]{|c|c|c|c|}
\hline
\textbf{Task} & \textbf{Physics Randomization Target} & \textbf{Range} & \textbf{Reward} \\ \hline
Walker Run & Contact Friction & {[}0.05 - 4.5{]} & Fixed \\ \hline
Pendulum Swingup & Mass Scaling Factor of Pendulum & {[}0.1 - 2.0{]} & Fixed \\ \hline
Throwing & Mass Scaling Factor of Ball & {[}0.2 - 1.0{]} & Fixed \\ \hline
Kuka Sorting & Mass Scaling Factor of Object & {[}0.2 - 1.0{]} & Parameterized \\ \hline
Pointmass & X/Y Motor Scaling Factor & \begin{tabular}[c]{@{}c@{}}X {[}1 - 2{]}\\ Y {[}1 - 2{]}\end{tabular} & Parameterized \\ \hline
\end{tabular}
\caption{Parameter randomization applied for each task.}
\label{table:tasks}
\end{table*}

% \begin{figure*}
% % \centering
%     \centering
%     \begin{tabular}{c c c c c}
%         \includegraphics[height=3cm,trim={0.2cm 0.2cm 14cm 0.25cm},clip]{images/Tasks.png} &
%         \includegraphics[height=3cm,trim={3.3cm 0.2cm 10.4cm 0.25cm},clip]{images/Tasks.png} &
%         \includegraphics[height=3cm,trim={6.7cm 0.2cm 6.8cm 0.25cm},clip]{images/Tasks.png} &
%         \includegraphics[height=3cm]{images/Kuka.png} & 
%         \includegraphics[height=3cm,trim={13.6cm 0.2cm 0.1cm 0.25cm},clip]{images/Tasks.png} \\
%         Walker & Pendulum & Throwing & Kuka Sorting & Pointmass \\
%     \end{tabular}
% \caption{Five HIP-MDP tasks used in our experiments.}
% \label{fig:tasks}
% \end{figure*}

When training the policy, we start with a seed state sampled from the replay buffer and then proceed in imagination only, as in the original DreamerV2. Via this setup, the actor and critic networks are trained to maximize the estimated discounted sum of rewards in imagination using a fixed world model.

% \begin{figure}
%     \centering
%     \vspace{-1em}
%     \includegraphics[width=0.5\textwidth]{images/Policy-Network.png}
%     \vspace{-1em}
%     \caption{Policy network training architecture. \sehoon{Can you check the resolution? It's a bit blurred.}\sehoon{Can you add a key message of this figure?}}
%     \label{fig:policy_model_architecture}
% \end{figure}

% \begin{wrapfigure}{r}{0.35\textwidth}
%     \centering
%     \vspace{-1em}
%     \includegraphics[width=0.35\textwidth]{images/Policy-Network.png}
%     \vspace{-1em}
%     \caption{Policy network training architecture. \sehoon{too small}}
%     \label{fig:policy_model_architecture}
% \end{wrapfigure}
However, the key difference from DreamerV2 is that both the actor and critic networks take the estimated parameter $\hat\omega_t$ from the prediction head as an additional input, as well as the reconstructed state $\hat{x}_t$. Because learning the parameter estimation is much faster than learning the world model, this new connection works almost the same as providing the ground-truth hidden parameter for the majority of the learning time. We will examine this behavior in the discussion section.
% For training the actor-critic network, we again start by sampling states $x_t$ from the environment. This is passed through a frozen world model where we train a critic $v_\psi$ to estimate the discounted sum of rewards and an actor $\pi_\theta$ to output actions that maximize the critic value. Training is done with imagined trajectories, improving sample efficiency as we do not need to interact with the environment. Our addition of the hidden parameter $\omega_t$ as input to the actor-critic helps improve reward over the default Dreamerv2 setup which only uses the latent variable as input. Additionally, we also see improvements by including the state $x_t$ as input, which we have without additional effort due to the original world model reconstruction loss.

\section{Experiments}

We evaluate \algname on several HIP-MDP problems to answer the following research questions:
\begin{enumerate}
    \item Can our \algname solve HIP-MDP problems more effectively than the baseline RL and domain adaptation algorithms?
    \item Can the estimation network accurately find ground-truth hidden parameters?
    \item What are the impacts of the HIP reconstruction loss and HIP-conditioned policy?
\end{enumerate}

% 1. Does adding a reconstruction loss for a hidden physics parameter $\omega$ improve a world model representation in environments with variable $\omega$?

% 2. How important is estimating $\omega$ and adding it as an input to the world model for accurately modeling the environment?

% 3. What is the impact of adding the state $x$ and $\omega$ along with the latent features $h$ and $z$ as inputs to the actor-critic networks?

% 4. How does our method compare to the baselines at operating in environments with physics variations?

% Tasks: Sorting, Throwing, DMC

% Models: DR, Ours, Ours no external LSTM, ours not using states/only features
% Models: Model-free, RMA?, some generalization model-based method?

\subsection{HIP-MDP Tasks}

We evaluate our model on a variety of continuous control tasks from the DeepMind Control (DMC) Suite~\cite{Tassa2018}, along with some tasks developed in MuJoCo~\cite{Todorov2012}. All tasks involve operating in a continuous control environment with varying physics. The tasks are as follows:
\begin{itemize}
    \item DMC Walker Run - Make the Walker run as fast as possible in 2D, where the contact friction is variable.

    \item DMC Pendulum Swingup - Swing a pendulum to an upright position, where the pendulum mass is variable.

    % \item DMC Cheetah Run - Make the Cheetah run as fast as possible in 2D, where the front and back motors are randomly scaled

    % Maybe need more detailed descriptions of these next two tasks
    % Rewards, image
    \item Throwing - Control a paddle to throw a ball into a goal, where the ball mass is variable.

    \item Kuka Sorting - Move an object to a desired location using a Kuka Iiwa manipulator arm, where the object mass is variable and the target location depends on the mass: heavier objects to the left and lighter objects to the right. The target trajectory is defined via the RL policy and is tracked by the Kuka arm using operational space control \cite{1087068}.

    \item DMC Pointmass - Move the point mass to the target location, where the x and y motors are randomly scaled. The target location depends on the motor scaling:  away from the center for high motor scaling and towards the center for lower motor scaling. 
\end{itemize}
When we design these tasks, we start by simply introducing randomization to the existing two tasks, DMC Walker Run and DMC Pendulum Swingup. Then, we purposely design the last two tasks, Kuka Sorting and DMC Pointmass, to incorporate a reward function that depends on their hidden parameters. Throwing also implicitly necessitates a policy for identifying the ball's mass and adjusting its trajectory. However, its reward function is not explicitly parameterized.

% \begin{figure*}
% % \centering
%     \centering
%     \begin{tabular}{c c c c c}
%         \includegraphics[height=3cm,trim={0.2cm 0 14cm 0},clip]{images/Tasks.png} &
%         \includegraphics[height=3cm,trim={3.3cm 0 10.4cm 0},clip]{images/Tasks.png} &
%         \includegraphics[height=3cm,trim={6.7cm 0 6.8cm 0},clip]{images/Tasks.png} &
%         \includegraphics[height=3cm,trim={10.3cm 0 3.55cm 0},clip]{images/Tasks.png} & 
%         \includegraphics[height=3cm,trim={13.6cm 0 0.1cm 0},clip]{images/Tasks.png} \\
%         Walker & Pendulum & Throwing & Sorting & Pointmass \\
%     \end{tabular}
% \caption{Five HIP-MDP tasks used in our experiments.}
% \label{fig:tasks}
% \end{figure*}

All the environments are visualized in Fig. \ref{fig:tasks} and their randomization ranges are summarized in Table~\ref{table:tasks}.

% Fix formatting

\subsection{Baseline Algorithms}

The baseline algorithms that we compare against are as follows:

% Should add simple names for ablations for easy reference in figure
\begin{itemize}
    \item DreamerV2 : original DreamerV2 model proposed by \cite{Hafner2020}.
    \item Proximal Policy Optimization (PPO): model-free, on-policy learning algorithm proposed by \cite{Schulman2017} using the implementation from \cite{stable-baselines3}.
    \item Soft Actor Critic (SAC): model-free, off-policy learning algorithm proposed by \cite{Haarnoja2018} using the implementation from \cite{pytorch_sac}.
    \item Rapid Motor Adaptation (RMA): model-free domain adaptation algorithm proposed by \cite{Kumar2021}, which estimates hidden parameters from a history of states and actions. We train an expert PPO policy with $\omega$ as input and compare to the student RMA policy, which is trained with supervised learning to match $\omega$ using a history of previous states.
    % \item PrivilegedDreamer without external estimation module (PrivilegedDreamer - Estimation): Ablate external LSTM to demonstrate the importance of this module
    % \item PrivilegedDreamer without external estimation module or additional policy inputs (PrivilegedDreamer - Estimation - Policy): Ablate both LSTM and privileged information given to policy to evaluate relevance of policy inputs
\end{itemize}

% trained in environment with domain randomization as specified in Table \ref{table:tasks}.
We selected our baselines to cover all the state-of-the-art in model-based/model-free, on-policy/off-policy, and domain randomization/adaptation algorithms. All models were trained for 2 million timesteps in each environment randomized as specified in Table~\ref{table:tasks}. 
% Comparisons between each model are based on the average reward in the training environment averaged over 100 runs. The results are shown in Table \ref{table:experiment_results}. Learning curves for all models are shown in Figure \ref{fig:learning_curves}. The means and standard deviation are computed over three random seeds.

\begin{table*}[]
\vspace{-1em}
\resizebox{\textwidth}{!}{%
\begin{tabular}{lcccccc}
\hline
\textbf{Method} & \multicolumn{1}{c}{\textbf{Walker}} & \multicolumn{1}{c}{\textbf{Pendulum}} & \multicolumn{1}{c}{\textbf{Throwing}} & \multicolumn{1}{c}{\textbf{Sorting}} & \multicolumn{1}{c}{\textbf{Pointmass}} & \textbf{Mean} \\ \hline
\multicolumn{1}{l|}{PrivilegedDreamer} & \multicolumn{1}{l|}{\textbf{766.20 $\pm$ 20.19}} & \multicolumn{1}{l|}{\textbf{563.14 $\pm$ 147.44}} & \multicolumn{1}{l|}{788.59 $\pm$ 45.66} & \multicolumn{1}{l|}{\textbf{554.65 $\pm$ 26.25}} & \multicolumn{1}{l|}{\textbf{670.23 $\pm$ 13.93}} & \textbf{668.56 $\pm$ 70.87} \\
\multicolumn{1}{l|}{DreamerV2 + Decoder + ConditionedNet} & \multicolumn{1}{l|}{576.89 $\pm$ 96.68} & \multicolumn{1}{l|}{329.80 $\pm$ 37.10} & \multicolumn{1}{l|}{785.78 $\pm$ 64.18} & \multicolumn{1}{l|}{180.85 $\pm$ 46.55} & \multicolumn{1}{l|}{492.77 $\pm$ 17.82} & 473.22 $\pm$ 58.87 \\
\multicolumn{1}{l|}{DreamerV2 + Decoder} & \multicolumn{1}{l|}{671.85 $\pm$ 10.46} & \multicolumn{1}{l|}{259.84 $\pm$ 26.08} & \multicolumn{1}{l|}{707.51 $\pm$ 20.63} & \multicolumn{1}{l|}{87.74 ~~$\pm$ 43.24} & \multicolumn{1}{l|}{480.96 $\pm$ 29.91} & 441.58 $\pm$ 28.21 \\
\multicolumn{1}{l|}{DreamerV2~\cite{Hafner2020}} & \multicolumn{1}{l|}{715.57 $\pm$ 39.95} & \multicolumn{1}{l|}{289.43 $\pm$ 214.12} & \multicolumn{1}{l|}{706.09 $\pm$ 26.24} & \multicolumn{1}{l|}{167.61 $\pm$ 33.38} & \multicolumn{1}{l|}{488.41 $\pm$ 3.60} & 473.42 $\pm$ 99.26 \\
\multicolumn{1}{l|}{SAC \cite{Haarnoja2018}} & \multicolumn{1}{l|}{475.22 $\pm$ 13.02} & \multicolumn{1}{l|}{454.67 $\pm$ 268.98} & \multicolumn{1}{l|}{\textbf{945.65 $\pm$ 17.02}} & \multicolumn{1}{l|}{74.85 ~~$\pm$ 88.03} & \multicolumn{1}{l|}{393.49 $\pm$ 210.47} & 468.78 $\pm$ 158.03 \\
\multicolumn{1}{l|}{PPO \cite{Schulman2017}} & \multicolumn{1}{l|}{79.73 ~~$\pm$ 10.95} & \multicolumn{1}{l|}{470.04 $\pm$ 324.05} & \multicolumn{1}{l|}{707.03 $\pm$ 115.63} & \multicolumn{1}{l|}{229.93 $\pm$ 181.12} & \multicolumn{1}{l|}{545.86 $\pm$ 72.22} & 406.52 $\pm$ 176.93 \\
\multicolumn{1}{l|}{RMA \cite{Kumar2021}} & \multicolumn{1}{l|}{75.28 ~~$\pm$ 11.31} & \multicolumn{1}{l|}{516.83 $\pm$ 386.43} & \multicolumn{1}{l|}{624.57 $\pm$ 118.70} & \multicolumn{1}{l|}{82.33 ~~$\pm$ 416.57} & \multicolumn{1}{l|}{545.31 $\pm$ 357.86} & 368.86 $\pm$ 305.00 \\\hline
\end{tabular}%
}
\caption{Model performance after 2 million timesteps of training. Our PrivilegedDreamer achieves better results particularly in the HIP-MDP problems with parameterized rewards, Kuka Sorting and Pointmass.
% \sehoon{Sort the order: Walker, Pendulum, Throwing, Sorting, and Pointmass.} \sehoon{Is PPO better than ours on Pointmass? It is against the learning curves.} \morgan{PPO differences are a result of smoothing the learning curves and using the raw value for the table. I changed it to use smoothing in the table as well.}
}
\label{table:experiment_results}
\end{table*}

\begin{figure}[t]
\vspace{-2em}
\begin{center}
\includegraphics[width=\linewidth]{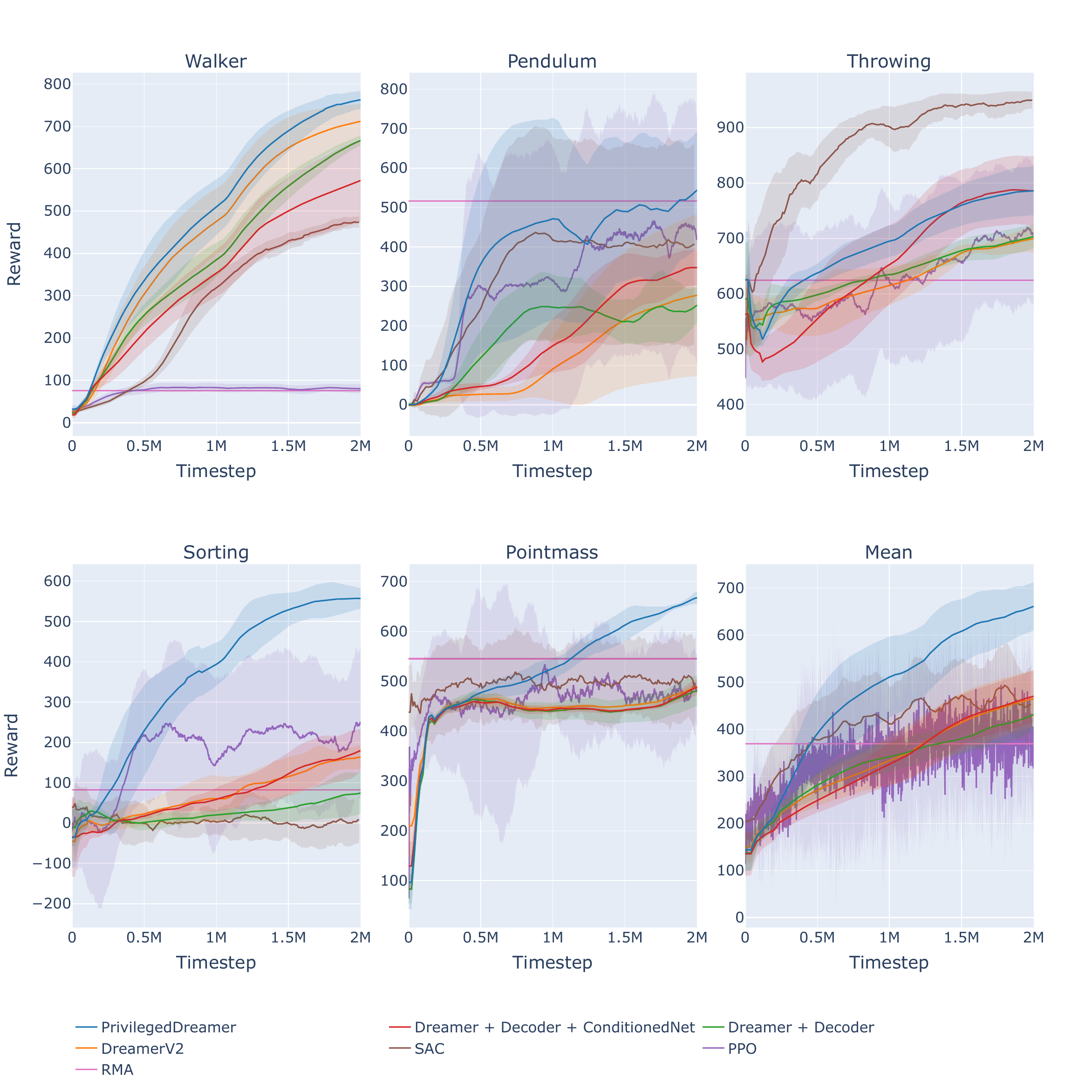}
\end{center}
\vspace{-1em}
\caption{Learning curves for all tasks. \algname shows the best performance against all the baseline algorithms, except for the throwing task which requires a very long horizon prediction.}
% \sehoon{Legends are too small. DR to DreamerV2. No RMA?} \sehoon{Can you update the color coding? Like, blue-ish color for our methods and other colors for baselines. Or line styles (dash/no-dash) may work.} \morgan{RMA is trained using an already trained Expert PPO policy. It's not trained for the same number of steps as the rest so isn't easy to compare. I can train it for longer if you think it should be included here}
% \sehoon{TODO: check RMA. If it takes longer, you can plot it as horizontal lines.}
% }
\label{fig:learning_curves}
\vspace{-1em}
\end{figure}

To validate our design choices, we further evaluate the following intermediate versions of the algorithm.

\begin{itemize}
    \item Dreamer + Decoder: This version only trains a decoder $\hat\omega_t \sim p_\phi(\hat\omega_t \vert h_t, z_t)$ by minimizing the hidden variable loss without an estimation module $\eta$. Also, $\hat\omega_t$ is not provided to the actor and critic and $h_t$ and $z_t$ are expected to contain all the information about the hidden parameter $\omega_t$. 
    \item Dreamer + Decoder + ConditionedNet: This version is similar to the previous Dreamer + Decoder, but the estimated $\hat\omega_t$ is given to the actor and critic networks. 
    % \item PrivilegedDreamer without external estimation module (PrivilegedDreamer - Estimation): Ablate external LSTM to demonstrate the importance of this module
    % \item PrivilegedDreamer without external estimation module or additional policy inputs (PrivilegedDreamer - Estimation - Policy): Ablate both LSTM and privileged information given to policy to evaluate relevance of policy inputs
\end{itemize}
Note that the proposed \algname can be viewed as the combination of Dreamer, an external estimation module, and conditioned networks trained with the hidden variable loss (\algname = Dreamer + ExternalEstimation + Decoder + ConditionedNet).

% Description of results
% Maybe include something showing ability to estimate actual value 
% versus other models and how that makes our work better

\subsection{Evaluation}

\paragraph{Performance}

To evaluate the effectiveness of the proposed method, we first compare the learning curves and the final performance of all the learned models. Learning curves for all models are shown in Figure~\ref{fig:learning_curves}, where the means and standard deviations are computed over three random seeds. Since RMA is trained in a supervised fashion using an expert policy and is not trained using on-policy environment interactions, we do not have a comparable learning curve, so we display the average performance as a horizontal line for comparison. Table~\ref{table:experiment_results} shows the average reward over $100$ runs for each seed. We also report the average performance over five tasks in both Figure~\ref{fig:learning_curves} and Table~\ref{table:experiment_results}.

% \begin{figure}[t]
% \vspace{-1em}
% \begin{center}
% \includegraphics[width=\linewidth]{images/Learning-Curves.pdf}
% \end{center}
% \vspace{-1em}
% \caption{Learning curves for all tasks. \algname shows the best performance against all the baseline algorithms, except for the throwing task.}
% % \sehoon{Legends are too small. DR to DreamerV2. No RMA?} \sehoon{Can you update the color coding? Like, blue-ish color for our methods and other colors for baselines. Or line styles (dash/no-dash) may work.} \morgan{RMA is trained using an already trained Expert PPO policy. It's not trained for the same number of steps as the rest so isn't easy to compare. I can train it for longer if you think it should be included here}
% % \sehoon{TODO: check RMA. If it takes longer, you can plot it as horizontal lines.}
% % }
% \label{fig:learning_curves}
% \vspace{-1em}
% \end{figure}

Overall, the proposed \algname achieves the best average reward over five tasks. It shows a significant performance improvement over the second best model, vanilla DreamerV2, in both standard DeepMind Control Suite tasks (Walker, Pointmass, Pendulum) as well as tasks we created ourselves (Sorting, Throwing). Performance margins are generally larger in the Sorting and DMC Pointmass tasks, where \algname is the \textbf{only model} tested that does appreciably better than random. This is likely because the reward for these tasks explicitly depends on $\omega$ and DreamerV2 only implicitly adapts its behaviors to the hidden parameters. This indicates that the novel architecture of \algname is effective for solving HIP-MDPs, particularly when the reward function is parameterized.
We suspect that RMA and PPO do especially poorly on the Walker task because the 2 million timestep training limit is insufficient for on-policy algorithms. Similarly, we suspect that the small training size affects the ability of RMA to effectively adapt, and that it would be more competitive with our method with a larger training dataset, which our method does not need due to its better sample efficiency.
% \sehoon{Morgan, provide more intuition/comment as much as possible. You can comment on each algorithm. RMA? PPO?}

\begin{figure}[t]
% \vspace{-2em}
\begin{center}
\includegraphics[width=\linewidth, trim={0 0 0 1.5cm},clip]{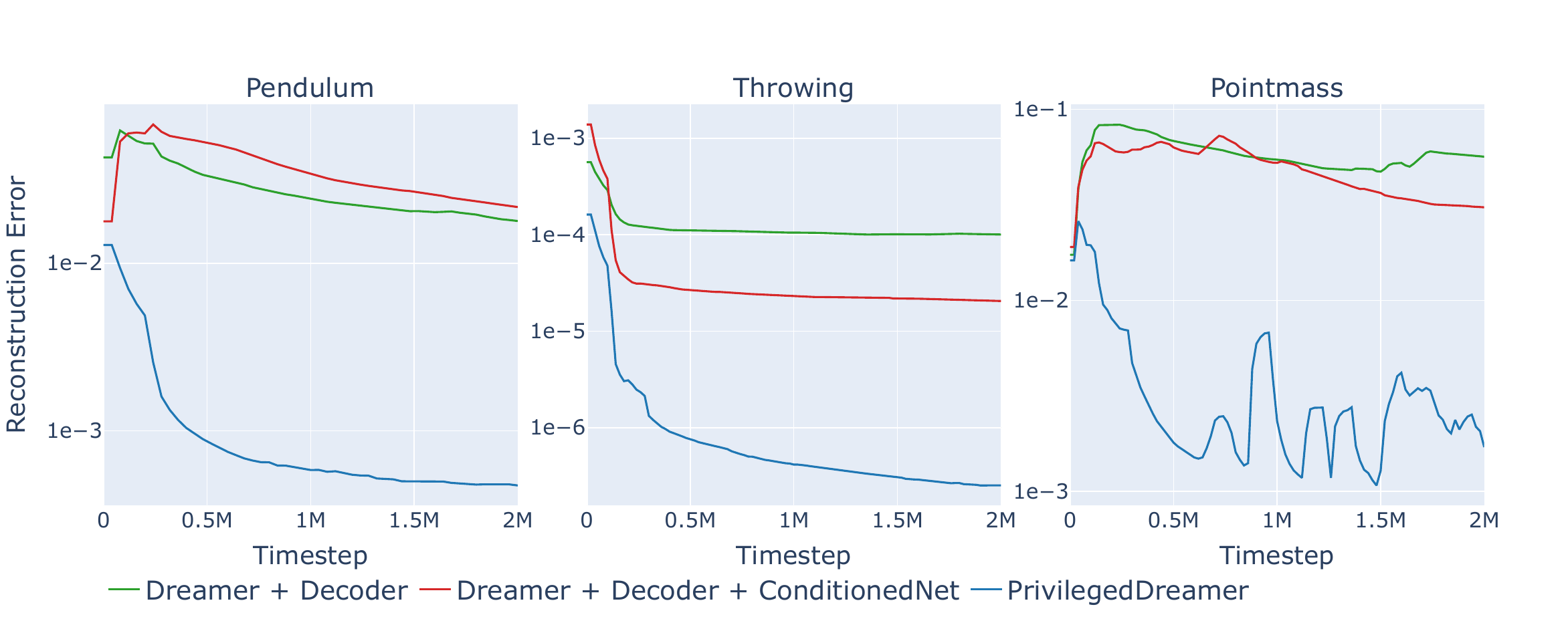}
\end{center}
\vspace{-1em}
\caption{Hidden parameter reconstruction error during learning. PrivilegedDreamer shows the best accuracy.}
% \sehoon{Looks nice. Can you remove some minor y-axis ticks for brevity? We can only keep 0.01, 0.001, and so on. I prefer the scientific notation, like $1e-5$.}
% \morgan{Not sure if this is useful. Some loss scaling differences between models makes this hard to interpret as just being the effect of each design choice.} \sehoon{Select Pendulum, Throwing, and Pointmass. Change the Y-axis to log-scale.}\sehoon{Can you also add a new figure to show online estimation within episodes for those three tasks? Please refer to the story in the paragraph ``Hidden Parameter Estimation''}
\label{fig:world_model_error}
\end{figure}

% \begin{figure}
% \vspace{-1em}
% \begin{center}
% \includegraphics[width=\linewidth,trim={0 0 0 1.5cm},clip]{images/Online_Estimation.pdf}
% \end{center}
% \vspace{-1em}
% \caption{Online parameter estimation within an episode. The two estimated values for the Pointmass model are shown in separate plots to improve readability.  
% % \morgan{Could use percent error rather than raw estimate if that is better.}
% \sehoon{Captions are too small}
% \sehoon{Summarzie a key message of these figures as a single line.}
%  \morgan{These figures are just referenced in the Hidden Parameter Estimation section to show that we can effectively estimate/reconstruct hidden parameters and in the ablation study as a way to demonstrate that the 0full method is necessary for good estimation/reconstruction. Is there something you want that isn't stated in these sections?}
% }
% \label{fig:online_estimation}
% \vspace{-1em}
% \end{figure}

One notable outlier is the great performance of SAC on the Throwing task. We suspect that the nature of the problem makes it difficult for model-based RL algorithms, both PrivilegedDreamer and DreamerV2.
Moreover, in this task, a policy only has a few steps to estimate its hidden parameters and predict the ball's trajectory, which can easily accumulate model errors over a long time horizon. On the other hand, SAC, a model-free RL algorithm, efficiently modifies its behaviors in a model-free fashion without estimating a ball trajectory. The on-policy model-free algorithms, PPO and RMA, are not sample-efficient enough to achieve good performance within two million steps.

% Comparisons between each model are based on the average reward in the training environment averaged over 100 runs. The results are shown in Table \ref{table:experiment_results}. Learning curves for all models are shown in Figure \ref{fig:learning_curves}. The means and standard deviation are computed over three random seeds.

% \paragraph{Effect of Hidden Variable Loss}

% For comparing the effectiveness of a world model representation, we focus on two main criteria: (1) accurately reconstructing the environment states, which is essential for imagination \morgan{Need to look at the data to see if there is a meaningful improvement with our method}, and (2) total reward of a policy trained using this representation. Given this, Figure \ref{fig:learning_curves} shows that just incorporating a reconstruction loss is insufficient for increasing reward and actually performs worse than the default Dreamerv2 architecture trained with domain randomization. 

\paragraph{Hidden Parameter Estimation}
\algname is based on the assumption that estimating hidden parameters is crucial for solving HIP-MDPs. Fig.~\ref{fig:world_model_error} and Fig.~\ref{fig:online_estimation} show the effectiveness of our model in reconstructing the hidden parameters and estimating them online.  Fig.~\ref{fig:world_model_error} illustrates the reconstruction errors during the learning process for the Pendulum, Throwing, and Pointmass tasks. In all cases, our \algname exhibits faster convergence, typically within less than $0.5$ million environmental steps, resulting in more consistent learning curves. Additionally, Fig.~\ref{fig:online_estimation} displays the real-time estimation of hidden parameters during episodes. Our model accurately predicts these parameters within just a few steps, enhancing the performance of the final policies. These findings justify the effectiveness of an external LSTM-based hidden-parameter estimation module.
\vspace{-0.125em}

\subsection{Ablation Studies}

% \paragraph{Benefit of External Estimation Module}
% This leads to including the external estimation module $\eta$ to more effectively estimate the hidden parameter $\omega$. With this, we see a much lower reconstruction error within our world model, shown in Figure \ref{fig:world_model_error}. This shows the importance of including $\omega$ as both an input to the world model encoder as well as the output from the world model decoder. This is further demonstrated by the fact that our full PrivilegedDreamer model outperforms all other baselines and ablations for all of our tasks other than the Throwing task, where the model-free SAC does better. Since this is a task we developed ourselves, we have no other direct results to compare against, but we hypothesize that SAC does better on this task due to long horizon world model errors. Since SAC is model-free, it trains with perfect observations from the actual environment, where our approach trains using imagined trajectories within the world model. This modeling error, along with the lack of ability to correct for errors after the ball is out of the reach of the throwing arm, makes it where SAC is better than all model-based methods, although our method is better than all other compared methods.

\begin{figure}
% \vspace{-0.em}
\begin{center}
\includegraphics[width=\linewidth,trim={0 0 0 1.5cm},clip]{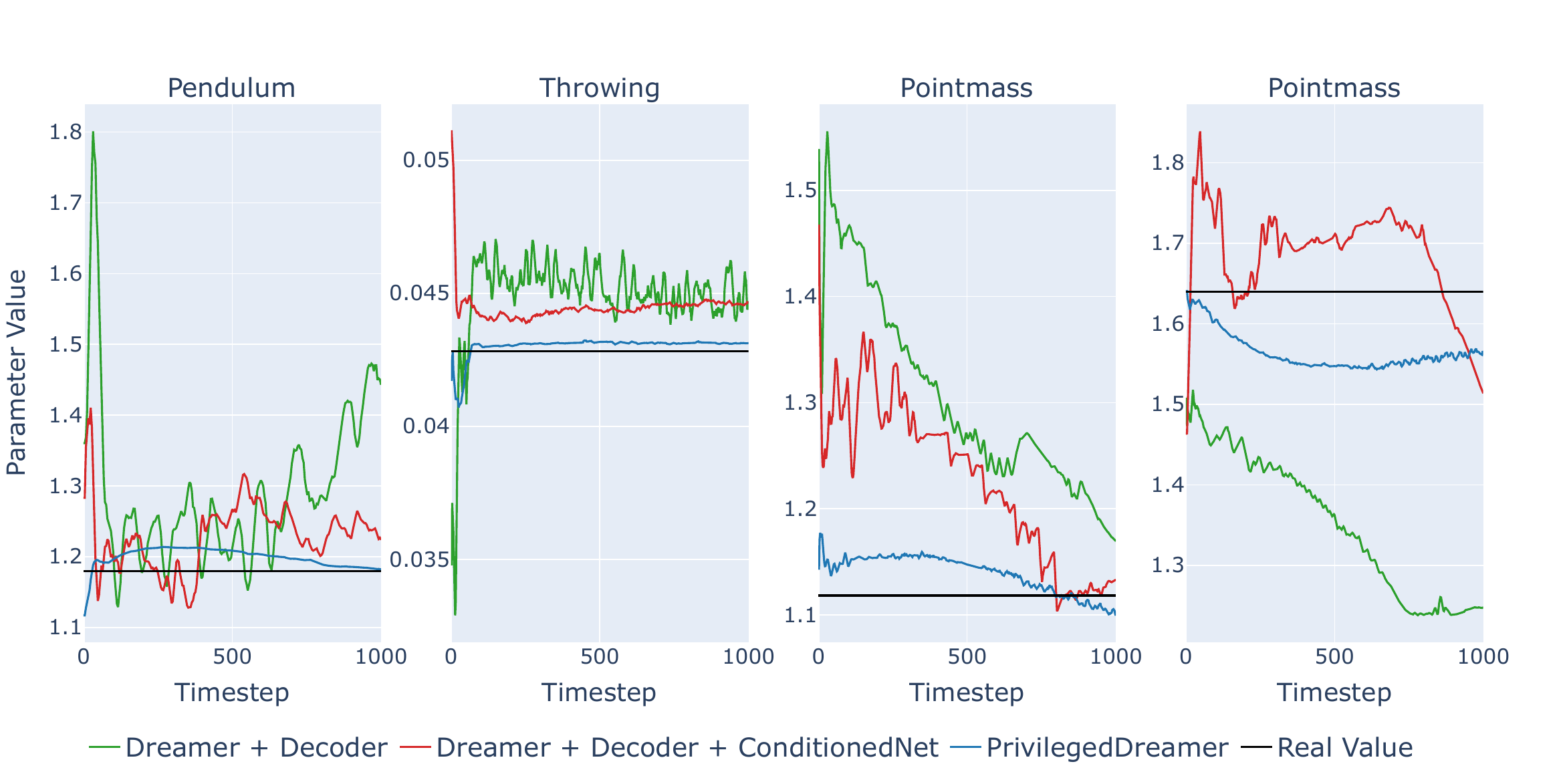}
\end{center}
\vspace{-1em}
\caption{Online parameter estimation within an episode. The two estimated values for the Pointmass model are shown in separate plots to improve readability. \algname was able to estimate hidden parameters more effectively than the other baselines.
% \morgan{Could use percent error rather than raw estimate if that is better.}
 % \morgan{These figures are just referenced in the Hidden Parameter Estimation section to show that we can effectively estimate/reconstruct hidden parameters and in the ablation study as a way to demonstrate that the full method is necessary for good estimation/reconstruction. Is there something you want that isn't stated in these sections?}
}
\label{fig:online_estimation}
\vspace{-1em}
\end{figure}

Comparing our full \algname model to the ablations, we see that our model is superior and each component is necessary for optimal performance. From Fig.~\ref{fig:world_model_error}, we see that our full model is significantly better at reconstructing the hidden variable $\omega$ than Dreamer + Decoder + ConditionedNet, which is already better than Dreamer + Decoder. With this low reconstruction error, online estimation of $\omega$ is very effective, as shown in Fig.~\ref{fig:online_estimation}, which shows that our method rapidly converges within 5\% of the real value, while the ablated versions take longer to converge to a lower quality estimate. Specifically, our agents find near-correct hidden parameters at the beginning of the episodes within a few environmental steps in all scenarios, while the other baselines take more than 500 steps (Dreamer+Decoder+ConditionedNet in Pointmass) or converge to wrong values (Dreamer+Decoder in Pendulum and Pointmass). Using this high quality estimate of $\omega$ within our ConditionedNet, Fig.~\ref{fig:learning_curves} and Table~\ref{table:experiment_results} demonstrate that our method greatly outperforms the ablations. This validates our hypothesis that incorporating a good estimate of $\omega$ into the world model and policy networks improves the performance of a RL policy operating in an environment with variable $\omega$.

% \sehoon{Can you write a new draft? For this time, I want you to focus on the comparison against Dreamer+Decoder and Dreamer+Decoder+ConditionedNet. Show that both an external estimation and conditioned networks are necessary. Walk over Figure~\ref{fig:learning_curves}, Table~\ref{table:experiment_results}, Figure~\ref{fig:world_model_error}, and a new figure to justify your claim.}

% \paragraph{Impact of Adding State and Hidden Variable to Policy Networks}

% Another idea we tested was to include the state $x$ and the hidden parameter $\omega$ as inputs to the policy and value networks along with the latent features $h$ and $z$. Comparing this to our full PrivilegedDreamer model and the other baselines, we see that this is better than naive DreamerV2 with domain randomization on some tasks and worse on others, with the mean being approximately equivalent to DR. This shows the benefit of this method over just adding a reconstruction loss, as the mean reward increased by 8\%. \sehoon{This is confusing. What did you compare exactly? The fair comparison would be \algname vs. \algname-state x.}
% \morgan{My goal was to attempt to show some benefit of just the ConditionedNet part by comparing it to DreamerV2. Not sure if we need this section at all or if the benefit of this component is adequately handled within other paragraphs.}

\section{Conclusion}
% \sehoon{Morgan, write down the draft here. I will follow up with you.}

This paper presents a novel architecture for solving problems where the dynamics are dictated by hidden parameters. We model these problems with  Hidden-parameter Markov decision processes (HIP-MDPs) and solve them using model-based reinforcement learning. We introduce a new model PrivilegedDreamer, based on the DreamerV2 world model, that handles the HIP-MDP problem via explicit prediction of these hidden variables. Our key invention consists of an external recurrent module to estimate these hidden variables to provide them as inputs to the world model itself. We evaluate our model on five HIP-MDP tasks, including both DeepMind Control Suite tasks and manually created tasks, where the reward explicitly depends on the hidden parameter. We find our model significantly outperforms the DreamerV2 model, as well as the other baselines we tested against. 
% Future topics of research for this work include applying this methodology to scenarios where the hidden parameter varies within an episode as well as using this for more practical, real-world applications, like robotics.

Our research opens up several intriguing agendas for future investigation. Firstly, this paper has concentrated our efforts on studying hidden parameter estimation within proprioceptive control problems, intentionally deferring the exploration of visual control problems like Atari games or vision-based robot control for future works. We believe that the same principle of explicitly modeling hidden parameters can be effectively applied to these visual control challenges with minor adjustments to the neural network architectures. Furthermore, we plan to investigate more complex robotic control problems, such as legged locomotion~\cite{Wu2022}, where real-world dynamics may be too sensitive to be precisely replicated by any of the hidden parameters. In such cases, we anticipate the need to devise better approximation methods.
Lastly, we plan to delve into multi-agent scenarios in which these hidden parameters have an impact on the AI behavior of other agents. These subsequent research directions hold promise in extending the scope and impact of the original paper.

% \newpage

\bibliographystyle{IEEEtran}
\bibliography{bib}

% Add appendices for hyperparameters, environment setups, etc.
%\newpage
%\input{appendix}

\end{document}